\pdfoutput=1
\documentclass[conference]{IEEEtran}
\IEEEoverridecommandlockouts

\usepackage{url}
\usepackage[T1]{fontenc}

\usepackage{cite}
\usepackage{amsmath,amssymb,amsfonts}
\usepackage{algorithmic}
\usepackage{graphicx}
\usepackage{textcomp}
\usepackage{xcolor}
\usepackage{multirow}
\usepackage{booktabs}
\usepackage{makecell}
\def\BibTeX{{\rm B\kern-.05em{\sc i\kern-.025em b}\kern-.08em
    T\kern-.1667em\lower.7ex\hbox{E}\kern-.125emX}}

\begin{document}

\title{ChestPheNoT: Deployable, Auditable Label--Status--Evidence
Extraction from Radiology Reports\\

\thanks{This work was supported by the National Institutes of Health's National Center for Complementary and Integrative Health under grant number R01AT009457, National Institute on Aging under grant number R01AG078154, and National Cancer Institute under grant number R01CA287413.}
}

\author{\IEEEauthorblockN{Kai Yu}
\IEEEauthorblockA{\textit{Department of Surgery} \\
\textit{University of Minnesota}\\
Minneapolis, USA \\
yu001014@umn.edu}
\and
\IEEEauthorblockN{Chenyu Zhu}
\IEEEauthorblockA{\textit{University of California Davis} \\
Davis, USA \\
cyuzhu@ucdavis.edu}
\and
\IEEEauthorblockN{Zaifu Zhan}
\IEEEauthorblockA{\textit{Department of Surgery} \\
\textit{University of Minnesota}\\
Minneapolis, USA \\
zhan8023@umn.edu}
\and
\IEEEauthorblockN{Meijia Song}
\IEEEauthorblockA{\textit{Department of Surgery} \\
\textit{University of Minnesota}\\
Minneapolis, USA \\
song0599@umn.edu}
\and
\IEEEauthorblockN{Min Zeng}
\IEEEauthorblockA{\textit{Department of Surgery} \\
\textit{University of Minnesota}\\
Minneapolis, USA \\
minzeng@umn.edu}
\and
\IEEEauthorblockN{Xiaoyi Chen}
\IEEEauthorblockA{\textit{Department of Surgery} \\
\textit{University of Minnesota}\\
Minneapolis, USA \\
chen9435@umn.edu}
\and
\IEEEauthorblockN{Mingquan Lin}
\IEEEauthorblockA{\textit{Department of Surgery} \\
\textit{University of Minnesota}\\
Minneapolis, USA \\
lin01231@umn.edu}
\and
\IEEEauthorblockN{Rui Zhang\textsuperscript{*}}
\IEEEauthorblockA{\textit{Department of Surgery} \\
\textit{University of Minnesota}\\
Minneapolis, USA \\
\textsuperscript{*}Corresponding author: ruizhang@umn.edu}
}

\IEEEpubid{\begin{minipage}{\textwidth}\ \\[10pt]\centering\footnotesize
\copyright~2026 IEEE. Personal use of this material is permitted. Permission from IEEE must be obtained for all other uses, in any current or future media, including\\
reprinting/republishing this material for advertising or promotional purposes, creating new collective works, for resale or redistribution to servers or lists,\\
or reuse of any copyrighted component of this work in other works.
\end{minipage}}

\maketitle

\begin{abstract}
Structured phenotype extraction from radiology reports supports cohort construction,
quality auditing, and clinical analytics, but practical deployment requires local
inference and auditable predictions, while expert annotations remain scarce.
Conventional labelers provide structured findings and assertion states
but no supporting evidence, while API-hosted large language models may be
unsuitable when clinical text cannot leave institutional infrastructure.
We present CHESTPHENOT, a compact 0.5--3B language model that jointly extracts
finding labels, three-class status (present/absent/uncertain), and verbatim supporting
evidence spans. CHESTPHENOT is trained using hybrid CheXbert+72B silver supervision
followed by supervised fine-tuning and lightweight GRPO refinement.
Across three human-annotated gold sets spanning in-distribution, cross-taxonomy,
and cross-institution evaluation, the 3B model remains below its CheXbert silver teacher
in distribution but is competitive under distribution shift, significantly surpassing CheXbert
on cross-institution detection (+2.0 F1). Task-specific training also enables the 3B model
to match or exceed substantially larger prompted models on most detection and status comparisons.
For evidence-grounded extraction, over $99\%$ of final evidence spans are locatable in the source
report, and the 3B model achieves 47.5 auditable-F1, outperforming Qwen2.5-7B one-shot
prompting by 7.6 points and approaching Qwen2.5-72B. These results demonstrate that
locally deployable models can provide competitive and directly auditable radiology-report
extraction without relying on external inference APIs. Code and the full
extraction/judge prompts will be made available at
\url{https://github.com/yukkai/ChestPheNoT}.
\end{abstract}

\begin{IEEEkeywords}
clinical natural language processing, radiology reports, phenotype
extraction, large language models, privacy-preserving deployment,
evidence grounding
\end{IEEEkeywords}

\section{Introduction}
\IEEEpubidadjcol
\label{sec:intro}

\begin{figure*}[t]
\centering
\includegraphics[width=\textwidth]{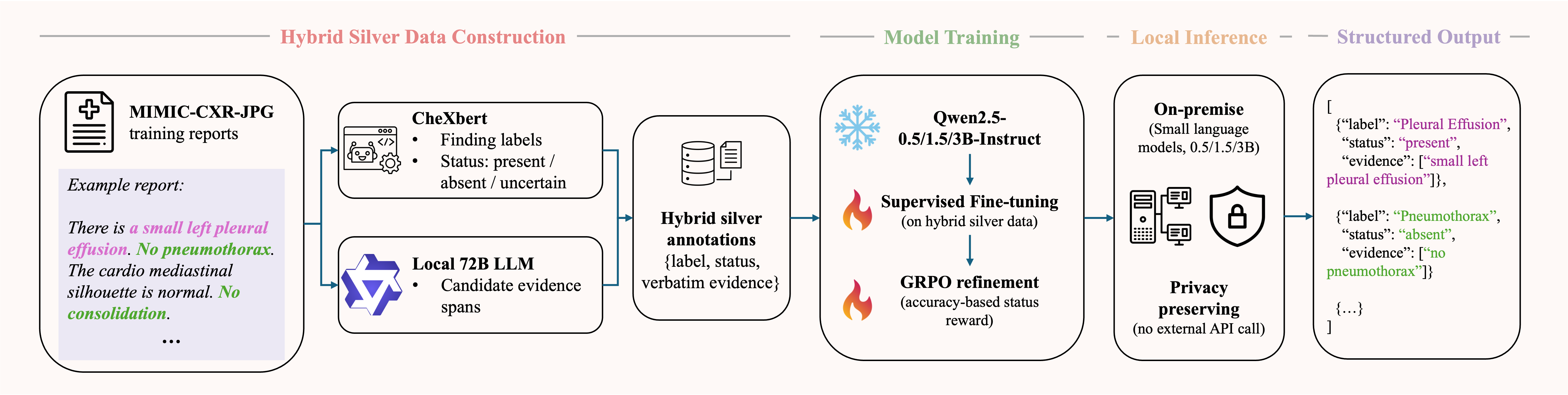}
\caption{Overview of \textsc{ChestPheNoT}. \emph{Hybrid silver
construction:} the CheXbert labeler supplies finding labels and status
(present/absent/uncertain) while a local 72B LLM proposes candidate
evidence spans, with exact-match filtering retaining only verbatim
report text, yielding \{label, status, evidence\} silver
annotations. \emph{Training:} a small Qwen2.5
model ($0.5$/$1.5$/$3$B) is supervised-fine-tuned on the silver data and
refined with GRPO. \emph{Inference:} the model runs on-premise with no
external API call, producing structured label/status/evidence output.}
\label{fig:overview}
\end{figure*}

Free-text radiology reports encode diagnostic findings needed for
cohort selection, registry population, quality assurance, and
clinical analytics, yet their unstructured form limits automated use.
Practical extraction systems face three constraints:
clinical text may be restricted from third-party APIs,
predictions should be traceable to supporting text,
and expert annotations are scarce. Conventional label/status systems
provide useful structured outputs but no supporting evidence,
while API-hosted large language models (LLMs) raise privacy,
cost, and grounding concerns. This leaves a gap between local deployment,
competitive extraction, and auditability.

We present \textsc{ChestPheNoT} (Fig.~\ref{fig:overview}), a compact
($0.5$--$3$B) Qwen2.5-based~\cite{qwen_qwen25_2025} extractor that
runs locally, extracts thirteen chest findings, and returns for each
identified finding a three-field record: a \emph{label}, a
\emph{status} (present/absent/uncertain), and one or more
\emph{verbatim evidence} spans copied from the report. The model is
trained by supervised fine-tuning on hybrid CheXbert$+$72B silver labels
and then refined with Group Relative Policy Optimization (GRPO)~\cite{shao2024deepseekmath}.

Our contributions are:
\begin{itemize}
\item A compact, locally deployable model that jointly predicts finding,
status, and explicit supporting evidence.
\item A label-scarce training recipe with evaluation across in-distribution,
cross-taxonomy, and cross-institution gold sets, using paired bootstrap
confidence intervals and comparisons against established labelers and
one-shot models up to 72B.
\item An evidence-quality evaluation that separates locatability from
semantic relevance and combines them with prediction correctness in
auditable-F1.
\end{itemize}

\section{Related Work}
\label{sec:related}

\textbf{Report labelers.}
CheXpert~\cite{irvin2019chexpert} and NegBio~\cite{peng2018negbio}
are widely used rule-based systems for extracting chest findings and
assertion status. CheXbert~\cite{smit2020combining} uses BERT with
rule-based and expert supervision, while CheX-GPT~\cite{gu2024chexgpt}
uses LLM-generated labels to train a lightweight BERT labeler.
These systems provide structured labels and assertion states but not
supporting evidence spans. MAPLEZ~\cite{lanfredi2025enhancing} produces
richer annotations with a locally executable 70B-class LLM and reports
CheXbert-comparable categorical F1 on the MIMIC/NIH annotations used
here, but under a different scoring protocol; its reported
$\approx$150\,GB VRAM footprint and two-A100 inference place it outside
our compact deployment tier, and it does not emit evidence spans.
RadGraph~\cite{jain2021radgraph} extracts span-level clinical entities,
relations, and certainty attributes, but does not normalize them to a
fixed finding taxonomy with per-finding status. \textsc{ChestPheNoT}
instead jointly predicts finding, status, and locatable evidence spans
within a fixed taxonomy.

\textbf{LLMs for clinical extraction.}
Generative LLMs support flexible structured extraction but may be
unsuitable when clinical text cannot be transmitted to external APIs.
Smaller task- or domain-specialized models can also match substantially
larger models on clinical NLP tasks~\cite{lehman2023need}. We therefore
fine-tune $0.5$B--$3$B models for local chest-report extraction with
explicit status and evidence output.

\textbf{Reinforcement learning for clinical LLMs.}
GRPO~\cite{shao2024deepseekmath} enables lightweight post-SFT
optimization. Prior work has applied GRPO to radiology-report disease
classification~\cite{wei2026reinforcement}, while
EvidenceRL~\cite{tamo2026evidencerl} and
HERO~\cite{zhao2026hero} optimize evidential reasoning.
Our GRPO reward uses label--status agreement and output validity, but
does not directly reward evidence quality, which is evaluated
independently.

\textbf{Evidence grounding and evaluation.}
Locatable evidence improves auditability but does not alone establish
semantic relevance~\cite{ji2023survey}. We therefore evaluate both
whether evidence is found in the source report and whether an
evidence-only LLM judge~\cite{zheng2023judging} considers it sufficient
for the predicted label--status claim. The judge is treated as a proxy
rather than clinical ground truth.

\section{Method}
\label{sec:method}

\subsection{Extraction Schema}
\label{sec:method-schema}
For a radiology report $x$, \textsc{ChestPheNoT} extracts findings from
the 13 standard CheXpert observations excluding the aggregate
\emph{No Finding}. For each identified finding $c\in\mathcal{C}$, the
model emits at most one record $(c,\hat{s}_c,\hat{E}_c)$, where
$\hat{s}_c\in\{\textrm{present},\textrm{absent},\textrm{uncertain}\}$
is its status and $\hat{E}_c$ contains supporting evidence spans.
Records are returned as a single JSON array, or an empty array when no
supported finding is identified. The intended contract requires
non-empty evidence copied from the source report for every emitted
record, making compliant predictions directly auditable.

\subsection{Hybrid Silver Data Construction}
\label{sec:method-silver}
We construct silver training data from 22k MIMIC-CXR radiology
reports~\cite{PhysioNet-mimic-cxr-2.1.0} sampled uniformly at random from
studies assigned to the recommended
training split distributed with MIMIC-CXR-JPG~v2.1.0~\cite{johnson2019mimic}.
No patient appears in both the silver-training corpus and the MIMIC gold evaluation sets.
CheXbert~\cite{smit2020combining} provides the finding labels and assertion
statuses. For each of the thirteen findings in $\mathcal{C}$, positive,
negative, and uncertain CheXbert outputs are mapped to \emph{present},
\emph{absent}, and \emph{uncertain}, respectively, while blank
(not-mentioned) findings are omitted. We use an on-premise
Qwen2.5-72B-Instruct model, solely during this
offline data-construction stage, to propose candidate evidence spans for
each non-blank CheXbert finding.
Candidate spans that are not exact substrings of the source report are
discarded, and a finding is retained only if at least one candidate span
survives this filter. Thus, although the 72B model may propose candidate
text that does not occur in the report, only verbatim substrings of the
report can enter the silver corpus. The resulting training examples follow the complete
$(c,s_c,E_c)$ schema without manual annotation.

\subsection{Model Training}
\label{sec:method-train}

We supervised-fine-tune Qwen2.5-Instruct models at $0.5$B, $1.5$B,
and $3$B scales on the silver corpus
(\textbf{SFT}), followed by Group Relative Policy Optimization
(GRPO)~\cite{shao2024deepseekmath} (\textbf{SFT+GRPO}).

For a sampled output $o=\{(\hat c_j,\hat s_j,\hat E_j)\}_{j=1}^{m}$ and
its silver target $y=\{(c_k,s_k,E_k)\}_{k=1}^{n}$, the reward observes
only the label--status pairs
$\hat P=\{(\hat c_j,\hat s_j)\}_{j=1}^{m}$ and
$T=\{(c_k,s_k)\}_{k=1}^{n}$, where $\hat P$ collects the pairs of all
valid records recoverable from $o$; a malformed output from which no
record can be recovered yields $\hat P=\emptyset$. We define
\begin{equation}
\mathrm{Match}(\hat P,T)=
\begin{cases}
1, & \hat P=T=\emptyset,\\[2pt]
\dfrac{1}{2}\left(\dfrac{|\hat P\cap T|}{|\hat P|}
+\dfrac{|\hat P\cap T|}{|T|}\right), & \hat P\neq\emptyset,\ T\neq\emptyset,\\[8pt]
0, & \text{otherwise},
\end{cases}
\label{eq:acc}
\end{equation}
the average of precision and recall with empty predictions and targets
handled explicitly, and the reward
\begin{equation}
R(o,y)=\lambda_{\mathrm{match}}\,
\underbrace{\mathrm{Match}(\hat P,T)}_{\text{label--status agreement}}
+\lambda_{\mathrm{fmt}}\,
\underbrace{\mathrm{Fmt}(o)}_{\text{structural validity}},
\label{eq:reward}
\end{equation}
where $\mathrm{Fmt}(o)=1$ if and only if $o$ parses as a JSON array in
which every record carries a label from the 13-class set, a status in
$\mathcal{S}$, and a non-empty evidence list. One exception prevents a
degenerate reward case: an output containing no finding record
(including an empty array) while the target is non-empty receives
$R=0$, with no format credit; this prevents structurally valid empty
predictions from collecting format-only reward whenever the target
contains findings.

At inference, we sample five generations per report; a label--status
pair is retained only if it appears in at least three of the five
generations. Evidence
is then aggregated across the agreeing generations. Each proposed span is
matched against the source report verbatim, allowing only whitespace
normalization, and successfully matched spans are expanded to their enclosing report sentences.
Unlocatable spans are discarded. The surviving sentences, which are therefore verbatim report
text by construction, are deduplicated and ordered by cross-generation frequency.
In the rare case where no evidence span for a majority-supported finding
can be located in the source report, the finding is emitted with its raw
span and flagged as non-locatable; contract violations are thus
retained and measured by the faithfulness metric rather than silently
enforced.

\section{Experimental Setup}
\label{sec:setup}

\subsection{Datasets}
\label{sec:setup-data}

We evaluate on three human gold sets spanning two taxonomies and two
institutions: \textbf{mimic-685}, 685 of 687 human-labeled MIMIC-CXR
test reports matched to report text in the CheXpert taxonomy
(in-distribution)~\cite{PhysioNet-mimic-cxr-2.1.0,johnson2019mimic};
\textbf{mimic-350}, 350 MIMIC-CXR reports with human MAPLEZ-taxonomy
annotations (cross-taxonomy; 62 reports overlap mimic-685); and
\textbf{nih-200}, 200 NIH reports with human MAPLEZ annotations
(cross-institution and cross-taxonomy)~\cite{lanfredi2025enhancing}.
For mimic-350/nih-200, predictions are mapped using the published
MAPLEZ merging rules: Lung Opacity absorbs its sub-findings and
Pneumonia is merged into Consolidation/Lung Opacity, leaving 12 scored
classes. Both MAPLEZ uncertainty types, including stability-related
uncertainty, map to \emph{uncertain}. mimic-685 is scored in the native
CheXpert taxonomy.

\subsection{Baselines and Metrics}
\label{sec:setup-metrics}

We compare with CheXpert~\cite{irvin2019chexpert},
NegBio~\cite{peng2018negbio}, and
CheXbert~\cite{smit2020combining}. Metrics are
\emph{present-only micro-F1} (13/12-class presence),
\emph{status macro-F1} over present/absent/uncertain, evidence
faithfulness (locatability) and relevance, and \emph{auditable-F1},
which credits a finding only when its status is correct and its evidence
is both locatable and judged relevant. Status is scored only on
gold-annotated cells. All generative models use five-sample per-finding
majority aggregation, and all one-shot baselines receive the same prompt
and worked example. We report 95\% percentile CIs from 1,000
report-level bootstrap resamples; pairwise comparisons use paired
bootstrap differences with two-sided $p$-values.

\subsection{Implementation Details}
\label{sec:setup-impl}

SFT uses full-parameter training for five epochs
($10^{-6}$ learning rate, cosine schedule, 0.03 warmup, effective batch
16, sequence length 2,048, bf16), with one run per configuration.
GRPO starts from the final SFT checkpoint and trains for one epoch at
the same learning rate, with
$\lambda_{\mathrm{match}}=0.7$, $\lambda_{\mathrm{fmt}}=0.15$,
KL coefficient 0.04, temperature 1.2, 16 generations per prompt,
completion length 512, and unscaled advantages
(Dr.\,GRPO~\cite{liu2025understanding}). Experiments ran on four
NVIDIA A100 80\,GB GPUs. We define the deployable ($\leq$8B) tier by
bf16 weight footprint: our $0.5$B--$3$B models require approximately
1--6\,GB and are compatible with a 24\,GB single-GPU memory budget,
whereas 70B-class models require multi-GPU serving.
Full configurations will be made available with the code.

\subsection{Evidence Relevance Judge and Data Use}
\label{sec:setup-judge}

Locatability does not establish semantic relevance, so we use
gpt-4.1-2025-04-14 (temperature 0) as an evidence-only judge. Given
only the predicted label, status, and cited evidence---not the full
report---it returns a binary judgment of whether the evidence supports
the claim. Because this requires a third-party API, relevance is
evaluated only on the publicly distributed, de-identified nih-200
reports. MIMIC-CXR remains entirely on-premise under its PhysioNet
data-use agreement, including silver-data construction with the 72B
teacher.

\section{Results}
\label{sec:results}

\subsection{Detection: In-Distribution vs.\ Out-of-Distribution}
\label{sec:res-present}
\begin{table}[t]
\centering
\caption{Present-only detection F1 and three-class status macro-F1
(\%) on mimic-685, mimic-350, and nih-200 (column headers
M-685/M-350/N-200). \textsc{ChestPheNoT} rows are SFT$+$GRPO.
\textbf{Bold}: best per column. $^\star$our silver label/status
teacher. $^\S$same base model as a trained row.}
\label{tab:main}
\footnotesize
\setlength{\tabcolsep}{3.2pt}
\begin{tabular}{@{}l ccc ccc@{}}
\toprule
& \multicolumn{3}{c}{Present F1} & \multicolumn{3}{c}{Status F1} \\
\cmidrule(lr){2-4}\cmidrule(lr){5-7}
Model & M-685 & M-350 & N-200 & M-685 & M-350 & N-200 \\
\midrule
\multicolumn{7}{@{}l}{\emph{\textsc{ChestPheNoT}}}\\
\quad 0.5B & 82.7 & 79.9 & 78.1 & 88.0 & 67.0 & 54.6 \\
\quad 1.5B & 84.0 & 79.5 & 81.1 & 89.3 & 66.2 & 53.8 \\
\quad 3B & 84.2 & \textbf{81.3} & \textbf{84.6} & 89.8 & \textbf{67.5} & \textbf{58.7} \\
\midrule
\multicolumn{7}{@{}l}{\emph{Established Labelers (no evidence output):}}\\
\quad CheXbert$^\star$ & \textbf{88.1} & 80.1 & 82.6 & \textbf{94.8} & 66.9 & 56.9 \\
\quad CheXpert & 83.6 & 70.9 & 78.6 & 78.2 & 43.3 & 49.5 \\
\quad NegBio & 84.4 & 71.1 & -- & 80.4 & 43.8 & -- \\
\midrule
\multicolumn{7}{@{}l}{\emph{One-shot Prompting:}}\\
\quad Llama-3.1-8B & 69.0 & 66.5 & 69.9 & 65.7 & 50.1 & 46.0 \\
\quad Qwen2.5-7B & 72.4 & 72.3 & 71.1 & 65.0 & 51.8 & 47.6 \\
\quad Qwen2.5-3B$^\S$ & 56.5 & 59.1 & 54.1 & 52.2 & 43.8 & 37.5 \\
\quad Llama-3.2-3B & 40.5 & 47.7 & 33.7 & 36.8 & 37.4 & 28.2 \\
\quad Qwen2.5-1.5B$^\S$ & 50.8 & 50.1 & 47.8 & 33.8 & 25.1 & 22.6 \\
\quad Llama-3.2-1B & 9.3 & 6.4 & 5.4 & 5.7 & 2.7 & 4.2 \\
\quad Qwen2.5-0.5B$^\S$ & 17.9 & 29.2 & 20.3 & 18.0 & 16.6 & 11.6 \\
\bottomrule
\end{tabular}
\end{table}
\begin{figure}[t]
\centering
\includegraphics[width=\columnwidth]{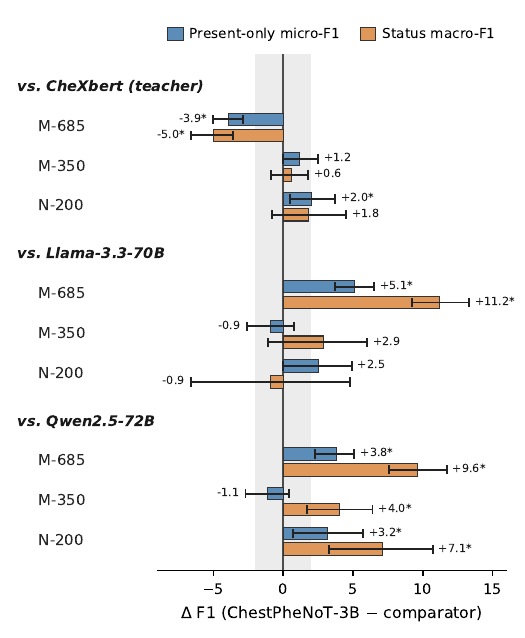}
\caption{\textbf{Paired $\Delta$ (ChestPheNoT-3B $-$ comparator)} on
present-only micro-F1 and status macro-F1, with $95\%$ paired
report-level bootstrap CIs; $\Delta>0$ favors our model. Shaded band:
$\pm2$\,F1, for scale. $*$: $p<0.05$.}
\label{fig:forest}
\end{figure}

Table~\ref{tab:main} reports present-only micro-F1, and
Fig.~\ref{fig:forest} shows the $3$B model's paired differences against
CheXbert, Llama-3.3-70B, and Qwen2.5-72B. Three patterns hold
across scales. First, our models outperform the rule-based labelers on
the cross-taxonomy set mimic-350 at every scale, by $+8.4$ to $+10.4$
present-F1 over CheXpert/NegBio (Table~\ref{tab:main}). Second,
against the strongest labeler CheXbert the in-distribution gap narrows
with scale and the model becomes competitive out of distribution: at
$0.5$B it is numerically comparable to CheXbert on mimic-350 ($79.9$
vs.\ $80.1$, Table~\ref{tab:main}); at $3$B it \emph{significantly surpasses}
CheXbert on nih-200, with a paired CI excluding zero ($+2.0$,
CI\,$[+0.5,+3.7]$; Fig.~\ref{fig:forest}), and is numerically ahead but
not significant on mimic-350 ($+1.2$, CI\,$[-0.0,+2.5]$;
Fig.~\ref{fig:forest}). Third, \emph{all} scales remain
below CheXbert on the in-distribution set mimic-685, with
the gap narrowing as size grows ($-5.4$ at $0.5$B to $-3.9$ at $3$B).
We return to the interpretation of the cross-taxonomy pattern in
Sec.~\ref{sec:discussion}.

\subsection{Three-Class Status Classification}
\label{sec:res-status}

\begin{figure}[t]
\centering
\includegraphics[width=\columnwidth]{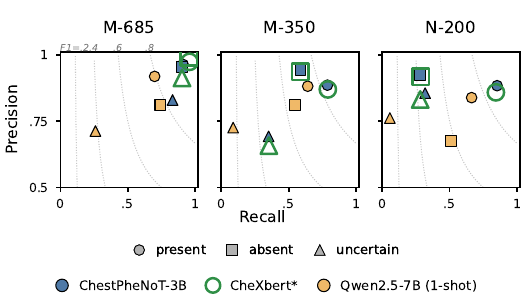}
\caption{\textbf{Per-status precision--recall} over gold-annotated
label cells (present/absent/uncertain support $=1465/660/316$ on
M-685, $1015/941/419$ on M-350, $439/354/280$ on N-200). Dotted:
iso-F1 contours. *our silver label/status teacher.}
\label{fig:pr}
\end{figure}

Table~\ref{tab:main} reports the three-class status macro-F1, and
Fig.~\ref{fig:pr} breaks it into per-status precision--recall for the
$3$B model, CheXbert, and the deployable one-shot baseline
(Qwen2.5-7B). Against the rule-based labelers the gap is large on every
set and scale ($+4$ to $+24$ macro-F1, Table~\ref{tab:main}), largest
on the cross-taxonomy set. Across the three sets, the \emph{uncertain}
class is the weakest axis for every system shown in Fig.~\ref{fig:pr}
and drives the out-of-distribution drop in macro-F1. Against CheXbert,
the teacher that supplied our label--status supervision, the trend
mirrors detection: below it in-distribution (narrowing from $88.0$ at
$0.5$B to $89.8$ at $3$B, vs.\ $94.8$; $3$B paired $\Delta{-}5.0$,
CI\,$[-6.6,-3.6]$, Fig.~\ref{fig:forest}) and not significantly
different from it out of distribution: $3$B$+$GRPO reaches
$67.5$ on mimic-350 ($\Delta{+}0.6$, CI\,$[-0.9,+1.8]$) and $58.7$ on
nih-200 ($\Delta{+}1.8$, CI\,$[-0.8,+4.5]$), vs.\ CheXbert's $66.9$ and
$56.9$ (Fig.~\ref{fig:forest}). The student thus shows no significant
status deficit to its teacher out of distribution, with point estimates
slightly ahead, while additionally producing auditable evidence spans,
a capability absent from the standard labelers
(Sec.~\ref{sec:res-evid}).

\begin{table}[t]
\centering
\caption{SFT vs.\ SFT$+$GRPO: present-only micro-F1 and status
macro-F1 (\%). The $+$GRPO rows equal the \textsc{ChestPheNoT} rows of
Table~\ref{tab:main}. Parentheses: change from SFT, positive in
\textbf{bold}.}
\label{tab:grpo}
\scriptsize
\setlength{\tabcolsep}{2.5pt}
\begin{tabular}{@{}l ccc ccc@{}}
\toprule
& \multicolumn{3}{c}{Present F1} & \multicolumn{3}{c}{Status F1} \\
\cmidrule(lr){2-4}\cmidrule(lr){5-7}
Model & M-685 & M-350 & N-200 & M-685 & M-350 & N-200 \\
\midrule
0.5B~~SFT     & 83.0 & 79.1 & 75.2 & 87.6 & 65.5 & 51.0 \\
\quad $+$GRPO & 82.7\,{\tiny($-$0.3)} & 79.9\,{\tiny\textbf{($+$0.8)}} & 78.1\,{\tiny\textbf{($+$2.9)}} & 88.0\,{\tiny\textbf{($+$0.4)}} & 67.0\,{\tiny\textbf{($+$1.5)}} & 54.6\,{\tiny\textbf{($+$3.6)}} \\
1.5B~~SFT     & 83.7 & 79.5 & 79.6 & 88.5 & 66.2 & 55.1 \\
\quad $+$GRPO & 84.0\,{\tiny\textbf{($+$0.3)}} & 79.5\,{\tiny($0.0$)} & 81.1\,{\tiny\textbf{($+$1.5)}} & 89.3\,{\tiny\textbf{($+$0.8)}} & 66.2\,{\tiny($0.0$)} & 53.8\,{\tiny($-$1.3)} \\
3B~~~~SFT     & 84.2 & 79.8 & 82.3 & 90.2 & 66.5 & 57.2 \\
\quad $+$GRPO & 84.2\,{\tiny($0.0$)} & 81.3\,{\tiny\textbf{($+$1.5)}} & 84.6\,{\tiny\textbf{($+$2.3)}} & 89.8\,{\tiny($-$0.4)} & 67.5\,{\tiny\textbf{($+$1.0)}} & 58.7\,{\tiny\textbf{($+$1.5)}} \\
\bottomrule
\end{tabular}

\vspace{2pt}
{\footnotesize $95\%$ paired bootstrap CIs for the $3$B SFT$\to$GRPO
step (present F1): $[-0.7,+0.6]$ on M-685, $[+0.5,+2.7]$ on M-350,
$[-0.1,+4.6]$ on N-200.}
\end{table}

\subsection{Effect of the Reinforcement-Learning Stage}
\label{sec:res-grpo}

Table~\ref{tab:grpo} isolates the RL stage. Its effect is small and
concentrated out of distribution: GRPO adds essentially nothing in
distribution (mimic-685 deltas between $-0.4$ and $+0.8$,
Table~\ref{tab:grpo}), while under shift the two sets behave
complementarily. On mimic-350 the $3$B step is significant in its own
right ($+1.5$, CI\,$[+0.5,+2.7]$, Table~\ref{tab:grpo}) though not
uniformly positive across scales ($0.0$ at $1.5$B); on nih-200 the
present-only step is positive at all three scales ($+1.5$ to $+2.9$;
one status cell negative at $1.5$B) but not individually significant
at $3$B (CI\,$[-0.1,+4.6]$), yet it suffices to turn a numerical tie
with CheXbert ($82.3$ vs.\ $82.6$, Tables~\ref{tab:grpo}
and~\ref{tab:main}) into a margin whose paired CI excludes zero
($+2.0$, CI\,$[+0.5,+3.7]$, Fig.~\ref{fig:forest}). These step-level
tests are exploratory, and each
configuration reflects a single training run. We therefore read GRPO as
an optional refinement concentrated under distribution shift rather
than a core contribution. Crucially, this gain does not appear to come
at the expense of evidence quality: on nih-200, the evidence-only judge
finds per-span relevance essentially unchanged from SFT to SFT$+$GRPO
(Sec.~\ref{sec:res-evid}).

\subsection{Trained Small vs.\ Prompted Large Models}
\label{sec:res-oneshot}

Table~\ref{tab:main} compares our trained models with one-shot
prompting of off-the-shelf instruction models ($0.5$--$8$B; the
$70$B-class ceiling references appear in Fig.~\ref{fig:forest}) under
the same schema prompt, with one worked example and no gradient
updates. Three findings stand out. First, one-shot prompting is
ineffective at small scale: one-shot Qwen2.5-0.5B reaches $18$--$29$
present-F1 and $12$--$18$ status macro-F1, versus $78$--$83$ and
$55$--$88$ for the trained model of the same size
(Table~\ref{tab:main}), a gain of up to $65$ present-F1 and $70$
status-F1 points from task-specific training. Second, the trained $3$B
model is competitive with the $70$B-class one-shot models on detection
despite having roughly $23\times$ fewer parameters: it exceeds both on
mimic-685 and nih-200 and trails them only on the cross-taxonomy
mimic-350 ($-0.9$/$-1.1$; Fig.~\ref{fig:forest}). Third, on three-class status the trained
$3$B model exceeds one-shot Qwen2.5-72B on \emph{all three} sets, with
paired CIs excluding zero ($+4.0$ to $+9.6$; Fig.~\ref{fig:forest}),
and exceeds Llama-3.3-70B in distribution ($+11.2$) while remaining
comparable to it on nih-200 ($-0.9$, CI\,$[-6.6,+4.8]$); for the
deployable one-shot baseline, the largest per-status deficit is on the
uncertain class (Fig.~\ref{fig:pr}). Overall, task-specific training
lets a compact model match or exceed substantially larger prompted
models while remaining in the deployable ($\leq$8B) tier.

\subsection{Evidence: Faithfulness and Auditability}
\label{sec:res-evid}
\begin{table}[t]
\centering
\caption{\textbf{Auditable-F1 on nih-200 (\%).} \textbf{Rec}: recall
over gold cells; \textbf{Prec}: precision over the system's own
predictions. \textbf{Bold} $=$ best in the deployable ($\leq$8B) tier;
\textcolor{gray}{gray} $=$ ceiling reference.}
\label{tab:evidence}
\footnotesize
\setlength{\tabcolsep}{12pt}
\begin{tabular}{@{}l rrr@{}}
\toprule
System & Rec & Prec & F1 \\
\midrule
\multicolumn{4}{@{}l}{\emph{Present} \ (gold cells $=439$):}\\
\quad \textsc{ChestPheNoT}-3B     & \textbf{81.1} & \textbf{84.2} & \textbf{82.6} \\
\quad Qwen2.5-3B         & 38.7 & 75.9 & 51.3 \\
\quad Qwen2.5-7B         & 62.0 & 78.4 & 69.2 \\
\quad \textcolor{gray}{Qwen2.5-72B} & \textcolor{gray}{77.7} & \textcolor{gray}{84.6} & \textcolor{gray}{81.0} \\

\midrule
\multicolumn{4}{@{}l}{\emph{Absent} \ (gold cells $=354$):}\\
\quad \textsc{ChestPheNoT}-3B     & 27.1 & \textbf{89.7} & \textbf{41.6} \\
\quad Qwen2.5-3B         & 23.2 & 31.4 & 26.7 \\
\quad Qwen2.5-7B         & \textbf{35.6} & 47.0 & 40.5 \\
\quad \textcolor{gray}{Qwen2.5-72B} & \textcolor{gray}{33.1} & \textcolor{gray}{68.4} & \textcolor{gray}{44.6} \\
\midrule
\multicolumn{4}{@{}l}{\emph{Uncertain} \ (gold cells $=280$):}\\
\quad \textsc{ChestPheNoT}-3B     & \textbf{12.5} & 33.7 & \textbf{18.2} \\
\quad Qwen2.5-3B         & 3.9 & 45.8 & 7.2 \\
\quad Qwen2.5-7B         & 5.4 & \textbf{71.4} & 10.0 \\
\quad \textcolor{gray}{Qwen2.5-72B} & \textcolor{gray}{11.1} & \textcolor{gray}{63.3} & \textcolor{gray}{18.9} \\
\midrule
\multicolumn{4}{@{}l}{\emph{Macro-F1:}}\\
\quad \textsc{ChestPheNoT}-3B     & & & \textbf{47.5} \\
\quad Qwen2.5-3B         & & & 28.4 \\
\quad Qwen2.5-7B         & & & 39.9 \\
\quad \textcolor{gray}{Qwen2.5-72B} & \textcolor{gray}{} & \textcolor{gray}{} & \textcolor{gray}{48.2} \\
\midrule
\quad \emph{Established labelers}$^{\ddagger}$ & n/a & n/a & n/a \\
\bottomrule
\end{tabular}

\vspace{2pt}
{\footnotesize $^{\ddagger}$ CheXbert, CheXpert, and NegBio do not
produce evidence spans and therefore are not auditable under this
metric.}
\end{table}

Every prediction carries an evidence span, so we ask two things: is it
\emph{locatable}, and does it \emph{support} the claim. Locatability is
learned rather than merely enforced: on the naturally formatted MIMIC
sets, $99.9\%$ of spans in the \emph{raw generations}, before any
aggregation or snapping, are already verbatim substrings of the report,
with unmatchable spans essentially absent ($\leq0.02\%$). On nih-200,
whose distributed reports are pre-tokenized (spacing around
punctuation, de-identification masks), exact string match drops for
formatting reasons ($95.9\%$ of the non-exact raw spans match exactly
after punctuation-spacing normalization), yet only $0.7\%$ of raw spans
cannot be aligned to the report, and after the standardization of
Sec.~\ref{sec:method-train} over $99\%$ of final spans are locatable.
The evidence-only judge of Sec.~\ref{sec:setup-judge} accepts
$89/96/42\%$ of final spans for present/absent/uncertain findings on
nih-200. Because a per-span acceptance rate is
not comparable across systems (its denominator is the system's own
output), we report \emph{auditable-F1} (Table~\ref{tab:evidence}): a
finding counts only if its status is correct, its span is locatable in
the report, and the judge accepts it; the locatability gate, applied
identically to all systems, prevents non-locatable evidence from
receiving credit from a report-blind judge.

Two results follow. \emph{First}, within the deployable ($\leq$8B)
tier, our $3$B model exceeds the deployable one-shot model,
Qwen2.5-7B, on all three statuses and by $+7.6$ auditable macro-F1
($47.5$ vs.\ $39.9$); at fixed size, training adds $+19.1$ over
one-shot prompting of the same base ($47.5$ vs.\ $28.4$; all
Table~\ref{tab:evidence}). \emph{Second}, against the ceiling reference
Qwen2.5-72B, roughly $23\times$ larger, it is within $0.7$ F1 points
($47.5$ vs.\ $48.2$), with a different operating profile: the $72$B
recovers more absent gold cells (Rec $33.1$ vs.\ $27.1$) at much lower
precision ($68.4$ vs.\ $89.7$, Table~\ref{tab:evidence}).
The established labelers emit
no evidence spans, so none of their records is auditable (n/a in
Table~\ref{tab:evidence}). Evidence relevance shows little change
under the label--status-only GRPO reward (Eq.~\eqref{eq:reward}):
per-span acceptance is $83.4$ vs.\ $82.4\%$ from SFT to SFT$+$GRPO, and
$85.0$ vs.\ $85.1$ on the $565$ findings emitted by both.

The weak point is \emph{uncertain} (F1 $18.2$; per-span acceptance
$42\%$; Table~\ref{tab:evidence}): the model tends to cite stability
language (``stable'', ``unchanged'') rather than explicit hedges, and
about two thirds of its uncertain evidence is of this kind. Notably,
the gold annotations themselves map stability-related uncertainty to
\emph{uncertain} (Sec.~\ref{sec:setup-data}), so citing stability
language is consistent with the gold convention, whereas the judge
prompt explicitly treats stability language as insufficient evidence of
uncertainty; the low acceptance on this class therefore partly reflects
a mismatch between the gold taxonomy and the operational definition
encoded in the judge rather than independent clinical adjudication.
Two corpus factors compound the difficulty: uncertain findings are the
rarest supervision signal ($9.8\%$ of the silver label--status
records), and their prevalence doubles from mimic-685 ($13\%$ of
gold-annotated cells) to nih-200 ($26\%$; Fig.~\ref{fig:pr}), which
depresses out-of-distribution macro-F1. The relevance axis also rests
on a single set ($n{=}200$) and a single LLM judge; radiologist
adjudication is left to future work.

\section{Discussion and Conclusion}
\label{sec:discussion}

\textsc{ChestPheNoT} demonstrates that compact, task-specific language
models can provide a practical alternative to conventional chest X-ray
labelers and substantially larger prompted models.
The $0.5$B--$3$B models support local inference while remaining
competitive with established labelers under distribution shift and producing
explicit supporting evidence. Together, the results establish a useful operating
point between predictive performance, deployment cost, and auditability.

The cross-taxonomy advantage partly reflects the MAPLEZ merge rather
than generalization: a one-shot Qwen2.5-72B with no task-specific
training also exceeds CheXbert on mimic-350 ($82.4$ vs.\ $80.1$), yet
under the same merge on nih-200 it trails CheXbert ($81.4$ vs.\
$82.6$) while our trained $3$B surpasses it ($+2.0$;
Fig.~\ref{fig:forest}); output-format flexibility alone therefore does
not explain the cross-institution gain.

Several limitations remain. First, all model scales remain below
CheXbert on in-distribution MIMIC, and uncertainty remains the weakest
status and evidence category. Second, evidence relevance is evaluated
with a single LLM judge and only on nih-200; independent radiologist
adjudication is needed. Finally, silver supervision inherits errors and
biases from CheXbert, which may limit further gains without stronger or
human-verified supervision.

Overall, these results support task-specific training of compact models
as a practical path toward locally deployable, evidence-grounded clinical NLP.

\bibliographystyle{IEEEtran}
\bibliography{refs}

\end{document}